\documentclass[10pt,twocolumn,letterpaper]{article}

\usepackage[pagenumbers]{cvpr} 

\usepackage{multirow}
\usepackage{listings}
\usepackage{graphicx}
\usepackage{booktabs}
\usepackage{pdflscape}
\usepackage{comment}
\usepackage[most]{tcolorbox}
\usepackage{subcaption}
\definecolor{cvprblue}{rgb}{0.21,0.49,0.74}
\usepackage[pagebackref,breaklinks,colorlinks,allcolors=cvprblue]{hyperref}

\def\paperID{*****} 
\def\confName{CVPR}
\def\confYear{2025}

\title{VISTA: \underline{V}ision-Language \underline{I}nference for Training-Free \underline{S}tock \underline{T}ime-Series \underline{A}nalysis}

\author{
Tina Khezresmaeilzadeh\textsuperscript{*}\\
University of Southern California\\
khezresm@usc.edu
\and
Parsa Razmara\textsuperscript{*}\\
University of Southern California\\
prazmara@usc.edu
\and
Seyedarmin Azizi\textsuperscript{*}\\
University of Southern California\\
seyedarm@usc.edu
\and
Mohammad Erfan Sadeghi\textsuperscript{*}\\
University of Southern California\\
sadeghim@usc.edu
\and
Erfan Baghaei Potraghloo\textsuperscript{*}\\
University of Southern California\\
baghaeip@usc.edu
}

\begin{document}
\maketitle
\renewcommand{\thefootnote}{\fnsymbol{footnote}}
\footnotetext[1]{All authors have contributed equally to this work.}

\renewcommand{\thefootnote}{\arabic{footnote}}
\setcounter{footnote}{0}
\begin{abstract}
Stock price prediction remains a complex and high-stakes task in financial analysis, traditionally addressed using statistical models or, more recently, language models. In this work, we introduce \textbf{\textit{VISTA}} (Vision-Language Inference for Stock Time-series Analysis), a novel, training-free framework that leverages Vision-Language Models (VLMs) for multi-modal stock forecasting. VISTA prompts a VLM with both textual representations of historical stock prices and their corresponding line charts to predict future price values. By combining numerical and visual modalities in a zero‑shot setting and using carefully designed chain‑of‑thought prompts, VISTA captures complementary patterns that unimodal approaches often miss. We benchmark VISTA against standard baselines, including ARIMA and text-only LLM-based prompting methods. Experimental results show that VISTA outperforms these baselines by up to 89.83\%, demonstrating the effectiveness of multi-modal inference for stock time-series analysis and highlighting the potential of VLMs in financial forecasting tasks without requiring task-specific training. The code is available at: \href{https://github.com/prazmara/MORFI}{Github link}

\end{abstract}    
\section{Introduction}
\begin{figure}[h]
    \centering
    \begin{subfigure}{0.45\textwidth}
        \centering
        \includegraphics[width=\linewidth]{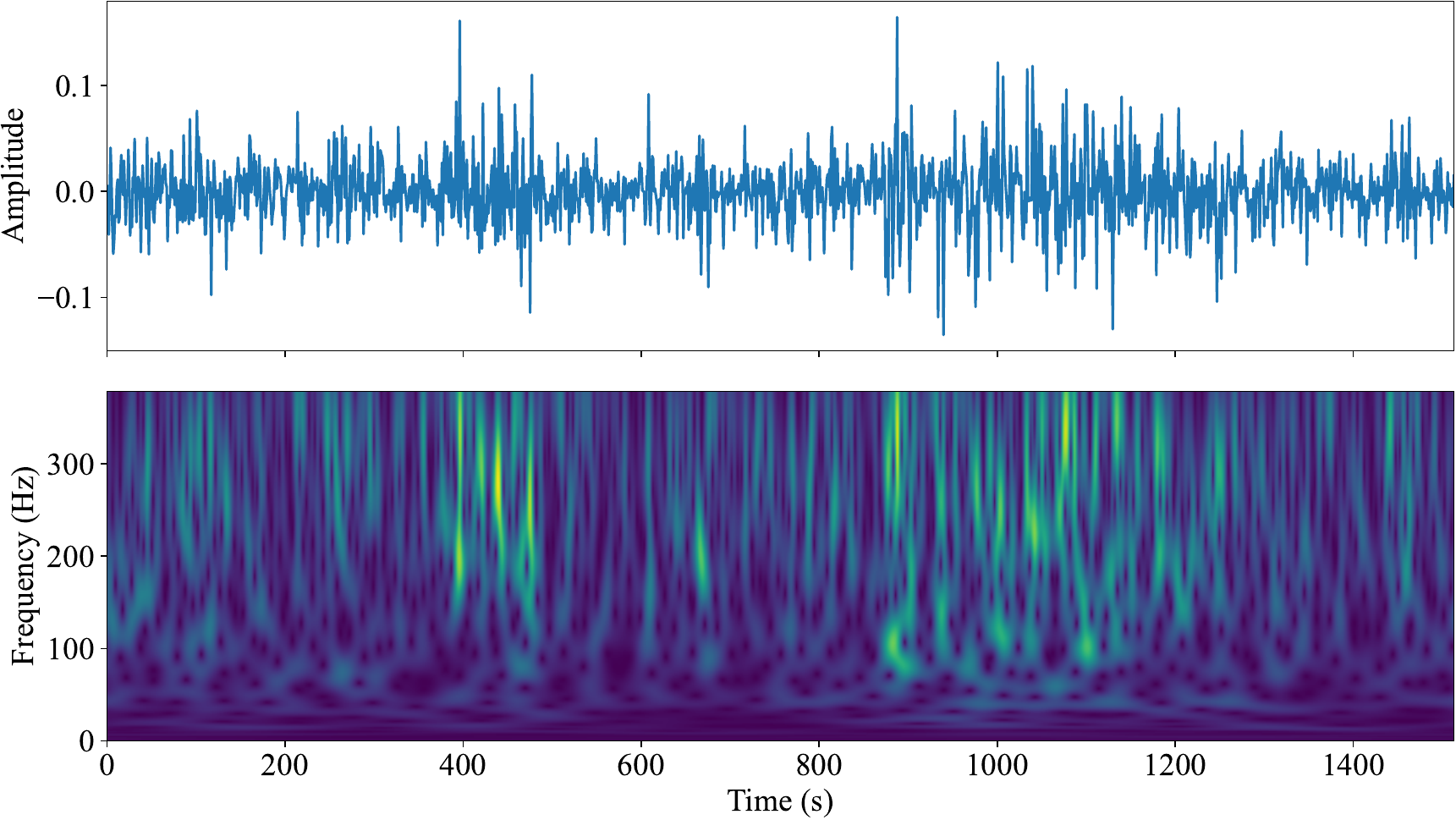}
        \caption{Accor stock value (first-order backward difference)}
        \label{fig:stockwell_a}
    \end{subfigure}

    \vspace{0.5em} 

    \begin{subfigure}{0.45\textwidth}
        \centering
        \includegraphics[width=\linewidth]{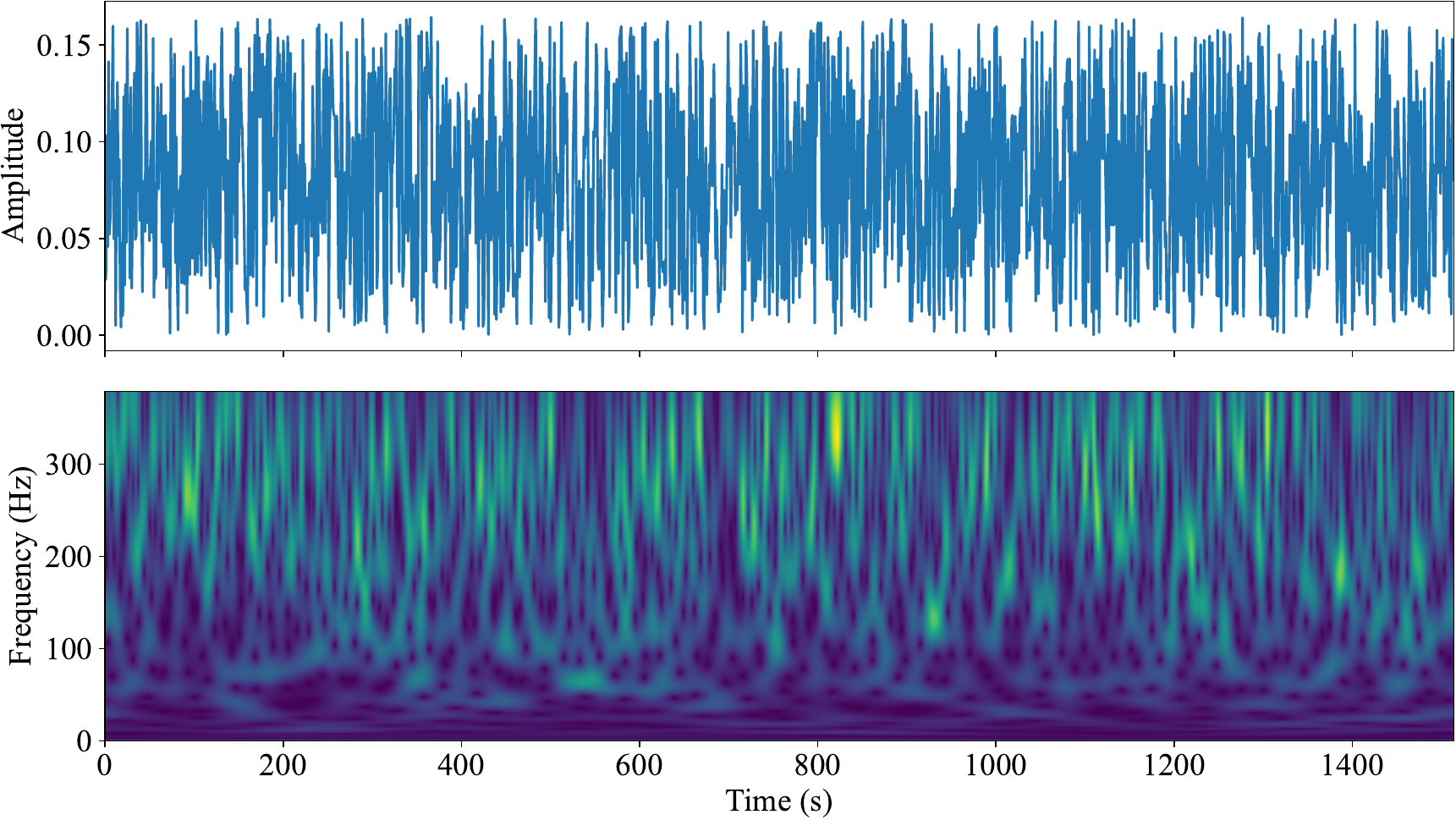}
        \caption{Uniform noise}
        \label{fig:stockwell_b}
    \end{subfigure}

    \caption{Comparison of Stockwell Transforms applied to Yahoo Finance Accor stock data and to synthetic noise.}
    \label{fig:stockwell_combined}
\end{figure}

Time-series forecasting plays a critical role in financial analysis, with stock price prediction standing out as a particularly impactful application. Accurate forecasting of stock prices can influence investment decision-making, risk assessment, and broader economic planning. Despite its importance, stock price prediction remains a complex challenge due to the high volatility and nonlinear dynamics inherent in financial markets. A significant portion of this complexity stems from the presence of noise—random, unpredictable fluctuations that obscure the underlying patterns of price movements. In fact, attempting to predict such noise is theoretically and practically infeasible. To illustrate this, a time-frequency analysis of the Accor Stock from \cite{yahoo_finance} and a synthetic unifomly distributed noise using the Stockwell transform\footnote{Stockwell is a time‑frequency technique that blends the windowed Fourier and wavelet approaches to yield a frequency‑dependent‑resolution spectrum while preserving absolute phase information.}\cite{stockwell1996localization} is presented in Fig. \ref{fig:stockwell_combined}. This visualization reveals a striking similarity between the transformed stock price signal and a corresponding noise signal, suggesting that a substantial component of the market behavior resembles random noise. Consequently, this highlights the intrinsic difficulty of price prediction, as it can be as challenging as forecasting the behavior of white noise itself. Traditional forecasting approaches often depend on large volumes of data and substantial computational resources, posing accessibility barriers for researchers with limited infrastructure or training capacity.

Recent advances in pre-trained foundation models—particularly large language models (LLMs) and vision-language models (VLMs)—open new possibilities for zero-shot and few-shot learning in stock price prediction. These models enable forecasting tasks to be performed without extensive fine-tuning or access to large training datasets. LLMs have demonstrated strong reasoning capabilities when prompted with structured inputs, while VLMs introduce the potential to incorporate visual representations of time-series data, such as plots, alongside textual context—supporting emerging multimodal approaches in financial forecasting.

Stock price prediction using foundation models poses several challenges, including the complexity and non-stationarity of financial time-series data, which are shaped by diverse factors such as economic indicators, market sentiment, and global events. These signals are often noisy and prone to abrupt fluctuations, making it difficult to extract consistent patterns—especially without domain-specific training or auxiliary information like news or financial reports.

Despite these limitations, developing resource-efficient, zero-shot forecasting approaches holds promise for democratizing access to predictive tools, enabling individuals and smaller institutions to benefit from financial insights without relying on large-scale training or computational infrastructure. Approaches that leverage structured prompts, visual data representations, or lightweight classical models may offer viable paths forward.

This work highlights the potential of pre-trained foundation models, particularly multimodal approaches, to use advanced financial forecasting tools while providing a resource-efficient solution for stock price prediction in real-world scenarios.

Specifically, we present \textit{\textbf{VISTA}}, a novel \textit{training‑free} approach in which both a line‑graph rendering of the time‑series and its underlying numerical values (expressed in textual form) are provided to a multimodal foundation model; the model’s generated text is then parsed to obtain the predicted stock prices. VISTA leverages carefully crafted chain‑of‑thought (CoT) \cite{wei2022chain} prompts that guide the vision‑language model through step‑by‑step reasoning about the series’ trend and seasonality, yielding more accurate forecasts.
\section{Related Works}

Over the last decade, machine learning has driven transformative progress in fields ranging from healthcare \cite{kolyshkina2021interpretability, razmara2024fever, ranjbar2025beyond, strzelecki2022machine} and recommendation systems \cite{khanal2020systematic, fayyazi2025facterfairnessawareconformalthresholding, khezresmaeilzadeh2025preserving, portugal2018use} to natural language processing \cite{vaswani2017attention, azizi2025mambaextend, aziziqmambaextend, abdollahi2025icd, gu2023mamba} and computer vision \cite{dosovitskiy2021imageworth16x16words, sadeghi2024peano, li2021survey} and privacy \cite{8406613, 10179281, khezresmaeilzadeh2024echoes, abbasi2024fedgreen, 9433648}. In financial settings, both classical ML time-series models (e.g., ARIMA, LSTMs) and more recent text-only LLM approaches have shown promise for short-horizon forecasting \cite{gogas2021machine}. Our work builds on these foundations by exploring whether adding visual trend information via vision-language models can further enhance performance without any task-specific training.

Stock forecasting has been approached using a wide variety of data modalities and model architectures. Recent progress in transformer-based models has shown notable performance on time-series data by modeling long-range dependencies through self-attention mechanisms. Notable examples include Informer \cite{zhou2021informer}, Temporal Fusion Transformers \cite{lim2021temporal}, and Autoformer \cite{wu2021autoformer}, all of which encode sequences as tokens and adapt transformer architectures to better handle temporal information.

An alternative line of research explores converting time-series data into images, enabling the use of vision transformers (ViTs). VisionTS \cite{chen2024visionts} introduced a visual masked autoencoder (MAE) framework that reconstructs missing parts of time-series representations in image form. This builds on prior work in masked modeling and self-supervised learning from computer vision, such as MAE \cite{he2022masked} and BEiT \cite{bao2021beit}. While VisionTS shows promising results in zero-shot and few-shot scenarios, it relies on periodicity in the time series for effective reconstruction—an assumption that does not always hold in financial data like stock prices, which often exhibit irregular, non-repeating trends \cite{achelis2000technical}.

Large Language Models, including BERT \cite{devlin2019bert} and GPT-style models, have demonstrated strong performance in zero-shot and few-shot tasks across domains through prompt-based learning. Their application to time-series forecasting—particularly financial data—is still emerging. LLMs can reason over historical stock prices via structured prompts, but their lack of built-in temporal modeling makes it difficult to capture complex dependencies without tailored prompt designs or architectural enhancements.

More recently, VLMs have enabled multimodal processing by jointly modeling visual and textual data. General-purpose models such as CLIP \cite{radford2021learning} and BLIP \cite{li2022blipbootstrappinglanguageimagepretraining} have shown success in tasks like image captioning and retrieval but are not directly designed for time-series forecasting. Models like DePlot \cite{liu2023deplotoneshotvisuallanguage}, which converts plots into tabular representations to be processed by LLMs, and PaliGemma \cite{beyer2024paligemma}, which integrates SigLIP \cite{zhai2023sigmoidlosslanguageimage} and Gemma \cite{gemmateam2024gemmaopenmodelsbased} for vision-language tasks, offer a more structured way to incorporate visual cues from data plots. However, these models still lack explicit mechanisms for modeling temporal dependencies, making them suboptimal for forecasting tasks without additional adaptation.

Our work builds upon these foundations by directly comparing LLMs and VLMs with aligned architectures and by investigating whether multimodal inputs (e.g., price plots) improve stock price forecasting. We also explore whether prompting strategies such as chain-of-thought reasoning can help bridge the gap in temporal understanding.
\section{Motivational Case Study}
In this section, we present a core motivation behind \textbf{VISTA}, our proposed vision-language paradigm for stock time-series forecasting. To frame the discussion, we begin with the following central question:

\begin{quote}
\textit{Why do we need a VLM that analyzes the graphical representation of a time series, if the same numerical values can be provided directly as text to a LLM?}
\end{quote}

This question addresses a fundamental limitation of uni-modal modeling: whether numerical sequences alone contain sufficient information for accurate forecasting, or whether visual representations can reveal complementary insights.

\subsection{Why Graphs Matter: Visual Representation Enhances Reasoning}

Human perception and reasoning are fundamentally multimodal. In neuroscience, studies show that the brain processes symbolic (numerical) and visual information through distinct but integrated pathways. The \textit{visual cortex} specializes in pattern recognition, while the \textit{intraparietal sulcus} and \textit{prefrontal cortex} support numerical and logical reasoning \cite{dehaene2003three, fuster2009cortex}.

Visual abstractions, such as line graphs and candlestick charts, enable the detection of emergent patterns—such as trends, cycles, or resistance levels—that are difficult to identify in numerical sequences alone \cite{kosslyn2006graph}. Traders and analysts rely on these representations to perform technical analysis, illustrating the practical and cognitive benefits of visual input.

This motivation aligns with findings in cognitive science, which suggest that humans are more accurate and efficient when combining visual and textual modalities \cite{tversky2002animation}. In AI, recent vision-language models like CLIP \cite{radford2021learning}, Flamingo \cite{alayrac2022flamingo}, and GPT-4V \cite{openai2023gpt4} demonstrate how multimodal learning can enhance generalization and contextual understanding beyond what is possible with unimodal input.

\subsection{Case Study: The Illusion of Stability}

Let us consider a concrete example that illustrates the limitation of using only text for time-series forecasting.

\paragraph{Time-Series Data:}
\[
[100, 102, 101, 100, 101, 102, 101, 100, 101, 100]
\]

\paragraph{Text-Only Prediction (LLM):}
Using only the raw numerical sequence, a Large Language Model such as Google Gemma predicts:
\[
[102, 101]
\]
This reflects a fluctuating trend without any inferred structural constraint—suggesting continued randomness.

\begin{figure}[h]
    \centering
    \includegraphics[width=0.45\textwidth]{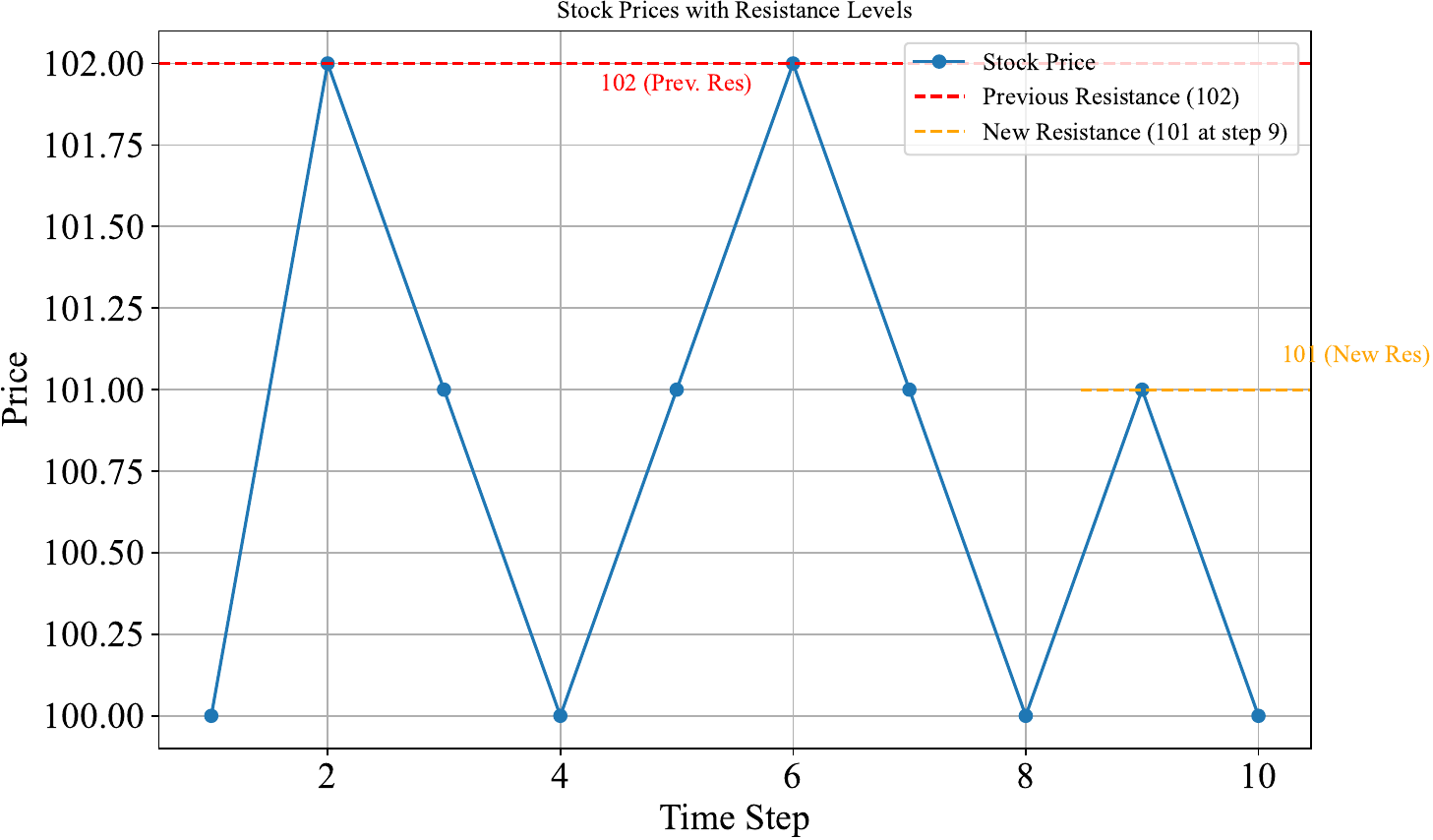}
    \caption{Graph of the 10-day time series showing a descending triangle pattern with resistance at 101.}
    \label{fig:triangle}
\end{figure}

\paragraph{Multimodal Prediction (VISTA):}
When the same model is prompted with both the time-series values and the corresponding line chart (see Figure~\ref{fig:triangle}), the prediction changes to:
\[
[101, 100]
\]
This indicates that the model recognizes the \textit{101 level} as a resistance—hit repeatedly but not broken—and expects a possible breakdown. The line chart clearly exhibits a \textit{descending triangle} formation, a common bearish indicator in technical analysis \cite{murphy1999technical}. This divergence in prediction shows that visual information provides structure and spatial cues that raw numbers do not. Without the graph, the model extrapolates fluctuations. With the graph, the model reasons about emerging patterns—demonstrating why multimodal input enables more robust inference in financial time-series forecasting.

\section{Methodology}
Motivated by the need for a multimodal architecture capable of zero-shot stock prediction, we provide a detailed overview of VISTA in this section.
\subsection{Problem Statement and Goal}


We address the task of short-term stock price forecasting given historical time-series data. Formally, let $\{p_1, p_2, \ldots, p_T\}$ denote the sequence of stock prices observed over $T$ consecutive time steps. The objective is to predict the future prices $\{\hat{p}_{T+1}, \hat{p}_{T+2}, \ldots, \hat{p}_{T+h}\}$ over a prediction horizon $h$.

In the \textit{text-only} setting, a Large Language Model is provided with the numerical values of the time series and tasked with generating the next $h$ predictions. In the \textit{multimodal} setting, a Vision-Language Model (VLM) receives both the numerical values and a line plot visualizing the same data. 

The main goal of this study is to investigate whether incorporating multimodal inputs improves the predictive performance of language models in time-series forecasting tasks. We aim to address the following research questions:

\begin{itemize}
    \item \textbf{Multimodal vs.\ Text-Only:} \textit{Does the inclusion of a visual representation of the time series (in the form of a line plot) enhance the forecasting ability of models when compared to text-only inputs?}
    
    \item \textbf{Chain-of-Thought Reasoning:} \textit{Can prompting models to explicitly describe their reasoning steps before making a prediction (i.e., Chain-of-Thought or CoT prompting) lead to more accurate and interpretable forecasts?}
\end{itemize}

By systematically evaluating these two strategies—multimodal input fusion and reasoning-based prompting—we seek to provide actionable insights into how generative models can be better leveraged for financial time-series prediction tasks.

\subsection{Multimodal Prompt Structure}
To evaluate whether VISTA offers improved predictive performance over text-only inputs in time series stock price prediction, we conducted a comparative analysis using VLMs and LLMs. Each VLM was paired with an LLM that shares a similar architectural structure (e.g., comparable backbone, number of parameters, and overall design). By aligning the architecture between the two, differences in performance can be attributed primarily to the inclusion of visual input.

We tested each model on identical stock price time series data but varied the way information was presented:

\begin{tcolorbox}[
  colback=gray!5!white,
  colframe=gray!80!black,
  title=Text-Only Prompt (LLM),
  fonttitle=\bfseries,
  boxsep=2pt,
  left=4pt,
  right=4pt,
  top=2pt,
  bottom=2pt,
  enhanced,
  sharp corners,
  before upper={\small}
]
\texttt{These are the time-series values of a stock over the first <PRICE\_LENGTH> days: <PRICE\_VALUES>.\\
Considering the time-series values, predict the stock price for the next <PREDICTION\_INTERVAL> days approximately.}
\end{tcolorbox}

\begin{tcolorbox}[
  colback=gray!5!white,
  colframe=gray!80!black,
  title=Multimodal Prompt (VLM),
  fonttitle=\bfseries,
  boxsep=2pt,
  left=4pt,
  right=4pt,
  top=2pt,
  bottom=2pt,
  enhanced,
  sharp corners,
  before upper={\small}
]
\texttt{This is the plot of a stock over the first <PRICE\_LENGTH> days, and these are the time-series values: \\
<PRICE\_VALUES>. Considering both the plot and time-series values, predict the stock price for the next <PREDICTION\_INTERVAL> days approximately.}
\end{tcolorbox}

In the multimodal setting, a line plot visualizing the historical stock prices over time was provided alongside the text prompt. This enabled the VLM to incorporate visual temporal patterns in addition to numerical sequences when generating predictions.

\subsection{CoT Solution Structure}
After our initial comparison between multimodal and text-only approaches, we explored whether Chain-of-Thought (CoT) prompting could further boost prediction accuracy. Chain-of-thought prompting is a method that enables complex reasoning by asking language models to generate intermediate reasoning steps that lead to the final answer \cite{wei2022chain}. We hypothesized that encouraging the model to break down its thinking process might not only improve clarity but also lead to more accurate and consistent forecasts.

To test this idea, we modified the prompt of VLM by adding a more detailed instruction. This instruction encouraged the model to first describe how it arrived at its prediction, and then provide the forecast.The following prompt was utilized to guide the model to include intermediate reasoning steps before determining the final values.

\begin{tcolorbox}[
  colback=gray!5!white,
  colframe=gray!80!black,
  title=CoT Prompt (VLM),
  fonttitle=\bfseries,
  boxsep=2pt,
  left=4pt,
  right=4pt,
  top=2pt,
  bottom=2pt,
  enhanced,
  sharp corners,
  before upper={\small}
]
\texttt{This is the plot of a stock over the first <PRICE\_LENGTH> days, and these are the time-series values: <PRICE\_VALUES>. Considering both the plot and time-series values, Examine if the trend is increasing, decreasing, stabilizing, or fluctuating.  predict the stock price for the next <PREDICTION\_INTERVAL> days approximately. This is a hypothetical projection based only on the trend in the graph and time-series values—ignore external factors like news or market sentiment. Only output the next <PREDICTION\_INTERVAL> predicted prices as a list.}
\end{tcolorbox}

\section{Evaluation and Results}
\subsection{Experimental Setup}
In this section, we outline the setup we used for the esperiments, including the datasets, models, evaluation metrics, and the text generation arguments. 

\subsubsection{Dataset and Preprocessing}
We use stock market data for a subset of companies from the CAC40 index, which includes the 40 largest publicly traded firms listed on the \textit{Euronext Paris} exchange \cite{euronext_paris}. In this work, we select four representative stocks: \textbf{Accor SA (AC.PA), BNP Paribas SA (BNP.PA), Capgemini SE (CAP.PA), and Air Liquide SA (AI.PA)}. These companies operate across sectors such as hospitality, banking, technology, and industrial gases.

Daily historical price data was collected from Yahoo Finance~\cite{yahoo_finance} for the period between \textit{January 1, 2014} and \textit{January 1, 2020}. The dataset includes values for \texttt{Open}, \texttt{High}, \texttt{Low}, \texttt{Close}, and \texttt{Volume}. For this study, we focus on the \texttt{Close} price. 
Each time series is transformed using Min-Max normalization, which maps the values to the interval $[0, 1]$ based on the following formula:
\[
x_{\text{scaled}} = \frac{x - x_{\min}}{x_{\max} - x_{\min}}
\]
This transformation adjusts the scale of the data while maintaining the original structure of price movements over time.

\subsubsection{Model Selection}

To isolate the contribution of visual context in multimodal forecasting, we formed five language–vision pairs in which the only substantive difference is the presence of a vision encoder and cross‑modal fusion layers. Each VLM inherits its language backbone, tokenizer, and most weights directly from the corresponding LLM, guaranteeing near‑identical parameter counts and transformer depth.

\textbf{T5-Base (220M) \cite{2020t5}}

\textbf{Google DePlot (282M) \cite{liu2022deplot}}
\begin{itemize}[leftmargin=1.5em]
    \item \textit{Reason for inclusion:} A token-based ``plot as pseudo-text'' model that treats plots as sequences of discrete tokens, enabling direct application of language models without adding vision encoders. This setup allows testing of multimodal capabilities with minimal deviation from a standard text-only transformer.
    \item \textit{Architectural parity:} DePlot builds directly on the T5-Base encoder-decoder architecture (220M parameters) \cite{raffel2020exploring}, adding only a lightweight plot-to-text parser and learned embeddings, resulting in a total of 282M parameters \cite{liu2022deplot}. The core transformer backbone remains unchanged, ensuring close architectural comparability.
\end{itemize}

\textbf{Llama-3.1-8B-Instruct \cite{meta_llama_3.1_8b_instruct_2024}} 

\textbf{LLaVA-1.5-7B-HF\cite{meta_llava-1.5-7b-hf_2024}}
\begin{itemize}[leftmargin=1.5em]
    \item \textit{Reason for inclusion:} A canonical open-source VLM that integrates vision via a CLIP encoder and maps it into the LLM embedding space using late-fusion cross-attention. Enables direct comparison between language-only and vision-augmented models with similar decoding stacks.
    \item \textit{Architectural parity:} LLaVA-1.5 uses a LLaMA-based 7B decoder \cite{touvron2023llama}, keeping the language core frozen. It integrates a CLIP ViT-L/14 vision tower (~428M parameters) and a lightweight projection MLP, resulting in a total of ~7.4B parameters—less than 6\% increase over the language-only baseline \cite{liu2023visual}.
\end{itemize}

\textbf{Phi-3-mini-128k-instruct(3.8B)\cite{microsoft_phi3_mini_128k_instruct_2024}}

\textbf{Phi-3-vision-128k-instruct(4.1B)\cite{microsoft_phi3_vision_128k_instruct_2024}}
\begin{itemize}[leftmargin=1.5em]
    \item \textit{Reason for inclusion:} A compact and instruction-tuned model family from Microsoft designed for high data quality and efficiency—ideal for evaluating multimodal gains in the low-parameter regime.
    \item \textit{Architectural parity:} Phi-3 Vision reuses the full Phi-3 Mini decoder-only language model (3.8B parameters) \cite{abdin2024phi} and adds a MiniCLIP vision encoder (~340M parameters) and a lightweight alignment layer, yielding a total of ~4.1B parameters. The core decoder remains unchanged for valid comparisons.
\end{itemize}

\textbf{Gemma-3-27B-IT LLM (27B) \cite{gemma_2025}}

\textbf{Gemma-3-27B-IT VLM (27B)\cite{gemma_2025}}
\begin{itemize}[leftmargin=1.5em]
    \item \textit{Reason for inclusion:} A state-of-the-art multilingual instruction-tuned model family with a unified architecture for both LLM and VLM variants. Evaluates whether large-scale vision-language integration improves over strong language-only baselines.
    \item \textit{Architectural parity:} Both models use the same decoder-only transformer architecture with 27B parameters \cite{team2024gemma}. The VLM variant of Gemma incorporates vision via a shared token space and cross-modal pretraining without modifying the core architecture. This ensures strict architectural comparability. \cite{gemma3_technical_report_2024}
\end{itemize}

\textbf{DeepSeek-R1-Distill-Qwen-1.5B(1.5B)\cite{deepseekai2025deepseekr1incentivizingreasoningcapability}}

\textbf{DeepSeek-VL-2-Tiny (3.37B) \cite{wu2024deepseekvl2mixtureofexpertsvisionlanguagemodels}}
\begin{itemize}[leftmargin=1.5em]
    \item \textit{Reason for inclusion:} DeepSeek-R1-Distill-Qwen-1.5B is a distilled, small-scale LLM tuned with reinforcement learning to improve mathematical and logical reasoning. Paired with DeepSeek-VL-2-Tiny, it allows us to isolate the effect of multimodal training in the low-parameter regime.
    \item \textit{Architectural parity:} Both models share the same Qwen2.5-1.5B decoder backbone \cite{guo2025deepseek}. The VL-2-Tiny variant prepends a Mixture-of-Experts (MoE) vision encoder (1.87B total, with 1B activated) and gating layers, bringing the full parameter count to 3.37B \cite{lu2024deepseek}. The language core remains intact, ensuring comparable language understanding capacity.
\end{itemize}

\subsubsection{Evaluation Metrics}
For each forecasting task, the model received the most recent 100 days of stock prices as input and was tasked with predicting the prices for the subsequent 5 days. Model performance was evaluated using four standard regression metrics:
\begin{itemize}
    \item Mean Squared Error (MSE)
    \item Root Mean Squared Error (RMSE)
    \item Mean Absolute Error (MAE)
    \item Mean Absolute Percentage Error (MAPE)
\end{itemize}
These metrics capture different aspects of prediction quality, with RMSE and MSE emphasizing larger deviations, while MAE and MAPE provide more interpretable, scale‐independent assessments of average error.


\subsection{Main Results}

\subsubsection{VLM-LLM Pairs}
\begin{table*}[]
\resizebox{\textwidth}{!}{%
\begin{tabular}{lllllllllllllllll}
\hline
\multirow{2}{*}{\textbf{LLM/VLM Pairs}} & \multicolumn{4}{c}{\textbf{Accor}}                          & \multicolumn{4}{c}{\textbf{Air Liquide}}                    & \multicolumn{4}{c}{\textbf{BNP Paribas}}                    & \multicolumn{4}{c}{\textbf{Capgemini}}                      \\ \cline{2-17} 
                                        & \textbf{MSE} & \textbf{RMSE} & \textbf{MAE} & \textbf{MAPE} & \textbf{MSE} & \textbf{RMSE} & \textbf{MAE} & \textbf{MAPE} & \textbf{MSE} & \textbf{RMSE} & \textbf{MAE} & \textbf{MAPE} & \textbf{MSE} & \textbf{RMSE} & \textbf{MAE} & \textbf{MAPE} \\ \hline
\textbf{T5}                             & 0.0213       & 0.1330        & 0.1223       & 34.6573       & \textbf{0.0071}       & \textbf{0.0646        }& \textbf{0.0593}       & \textbf{22.9027}       & 0.0326       & 0.1347        & 0.1174       & 42.6376       & 0.0206       & 0.0942        & \textbf{0.0728}       & 49.3178       \\
\textbf{DePlot}                  & \textbf{0.0181}       & \textbf{0.1214 }       & \textbf{0.1148 }      & \textbf{33.1477}       & 0.0137       & 0.0924        & 0.0896       & 41.7180       &\textbf{ 0.0164 }      & \textbf{0.1131}        & \textbf{0.1042}       & \textbf{24.1590}       & \textbf{0.0115}       & \textbf{0.0780}        & 0.0737       & \textbf{60.7436}       \\ \hline
\textbf{Llama3}                         & 0.0413        & 0.1955         & 0.1828        & 78.7101         & 0.0138        & 0.1083         & 0.1050        & 26.5555         & 0.0397        & 0.1688         & 0.1652        & 23.8431         & 0.0054        & 0.0705         & 0.0655        & 10.0217         \\
\textbf{LLaVA}                          & \textbf{0.0046}        & \textbf{0.0618 }        & \textbf{0.0481 }       & \textbf{16.5818 }        & \textbf{0.0010 }       & \textbf{0.0288}         & \textbf{0.0265}        & \textbf{6.4364 }        & \textbf{0.0072  }      & \textbf{0.0770  }       & \textbf{0.0711}        & \textbf{11.0233 }        & \textbf{0.0015 }       & \textbf{0.0362 }        & \textbf{0.0293}        & \textbf{3.8947 }        \\ \hline
\textbf{Gemma-LLM}                         & 0.0098        & 0.0864         & 0.0780        & 20.5006         & 0.0043        & 0.0585         & 0.0532        & 16.3828         & 0.0045        & 0.0639         & 0.0578        & 15.0268         & 0.0017        & 0.0369         & 0.0339        & 22.0976         \\
\textbf{Gemma-VLM}                       &  \textbf{0.0058}       &  \textbf{0.0644}        & \textbf{0.0590}       & \textbf{15.4665       }&  \textbf{0.0031}      &  \textbf{0.0467}        & \textbf{0.0424}      & \textbf{11.9728}     &  \textbf{0.0041}       &  \textbf{0.0561}       &  \textbf{0.0491}    &  \textbf{10.6249}       &  \textbf{0.0007}      &  \textbf{0.0232}       &  \textbf{0.0207}      &  \textbf{17.4091 }    \\ \hline
\textbf{Phi3}                           & 0.0459        & 0.1925         & 0.1892        & 41.8192         &   0.0172         &  0.1046          &0.1036     &  38.6716         &  0.0367         & 0.1601          & 0.1545        & 30.7799       & 0.0177       & 0.1257          & 0.1231         & 73.4928         \\
\textbf{Phi3Vision}                     & \textbf{0.0095}        & \textbf{0.0809}         & \textbf{0.0752 }       & \textbf{13.9770}         &\textbf{0.0014}        & \textbf{0.0331}          & \textbf{0.0306}       & \textbf{12.1948}         &  \textbf{0.0060}        &  \textbf{0.0713}          &  \textbf{0.0659}    & \textbf{16.1038}        & \textbf{0.0018}       & \textbf{0.0335}        & \textbf{0.0305}        & \textbf{19.0512}        \\ \hline
\textbf{DeepSeek-R1}                    & 0.0608       & 0.1885        & 0.1862       & 85.2315         & 0.0186       & 0.1065        & 0.1030       & 49.8984         & 0.0243       & 0.1275        & 0.1228       & \textbf{30.6346}         & 0.0206      & 0.1032        & 0.0981       & 31.0955        \\
\textbf{Deepseek\_VL2}                  & \textbf{0.0180}       & \textbf{0.1111}        & \textbf{0.1059}       & \textbf{26.4934}       & \textbf{0.0181}       & \textbf{0.1012}        & \textbf{0.0979}       & \textbf{45.0789}       & \textbf{0.0219}       & \textbf{0.1191}        & \textbf{0.1137}       & 39.6341       & \textbf{0.0083}       & \textbf{0.0719}        & \textbf{0.0674}       & \textbf{20.2569}       \\
 \hline
\end{tabular}%
}
\caption{Evaluation of various LLM/VLM pairs (e.g., T5, Google DePlot, LLaMA3, LLaVA, Phi, DeepSeek) on stock price prediction across four companies (Accor, Air Liquide, BNP Paribas, and Capgemini). Performance metrics include MSE, RMSE, MAE, and MAPE (expressed as \%)}
\label{tab:my-table}
\end{table*}
To assess the benefits of multimodal learning for stock price prediction, we compare VLMs and their language-only counterparts (LLMs) across the mentioned four companies: Accor, Air Liquide, BNP Paribas, and Capgemini and evaluate them using MSE, RMSE, MAE and MAPE.

Starting with T5 and its VLM variant DePlot, DePlot achieves better results on three out of four companies. For example, on BNP Paribas, DePlot lowers the MSE from 0.0326 to 0.0164, a 49.7\% reduction, and on Capgemini from 0.0206 to 0.0115 (44.17\% improvement). The only exception is Air Liquide, where T5 slightly outperforms DePlot (MSE of 0.0071 vs. 0.0137), but overall, the trend favors the multimodal model.

In the case of LLaMA-3 and LLaVA, the addition of visual input leads to notable performance improvements across all companies. On Accor, LLaVA achieves an MSE of 0.0046, compared to LLaMA-3’s 0.0413—a dramatic 88.9\% improvement. Similarly, on Capgemini, the MSE drops from 0.0054 to 0.0015, marking an 72.22\% reduction. These results highlight the strong impact of visual context in time-series forecasting.
The Gemma model pair also shows consistent improvement. On Accor, Gemma-VLM reduces the MSE from 0.0098 to 0.0058 (40.8\% improvement), and on Capgemini from from 0.0017 to 0.0007 (58.82\% improvement), confirming the effectiveness of multimodal inputs.
A clear trend is also observed in the Phi3 pair. Phi3Vision noticably outperforms Phi3 across the board, with the best improvement on Accor—reducing the MSE from 0.0459 to 0.0095, a remarkable 79.3\% decrease. On Capgemini, it improves from 0.0177 to 0.0018 (89.83\% reduction), again showcasing the value of incorporating visual cues.
Finally, DeepSeek-VL2 also shows improved performance over DeepSeek-R1. On Accor, the MSE drops from 0.0608 to 0.0180, and on BNP Paribas from 0.0243 to 0.0219. Though the gains are slightly more modest compared to other pairs, the trend still favors the multimodal variant.

In summary, across all model comparisons and companies, Vision-Language Models outperform their language-only counterparts. The consistent reduction in MSE—often by 40–80\%—confirms that the inclusion of visual input in the form of line plots significantly enhances the forecasting accuracy of generative models in stock price prediction tasks.


\subsubsection{VLM CoT over not CoT}
\begin{table*}[]
\resizebox{\textwidth}{!}{%
\begin{tabular}{llllllllllllllllll}
\hline
\multirow{2}{*}{\textbf{Model}}        & \multirow{2}{*}{\textbf{Prompting}} & \multicolumn{4}{c}{\textbf{Accor}}                          & \multicolumn{4}{c}{\textbf{Air Liquide}}                    & \multicolumn{4}{c}{\textbf{BNP Paribas}}                    & \multicolumn{4}{c}{\textbf{Capgemini}}                      \\ \cline{3-18} 
                                            &                                            & \textbf{MSE} & \textbf{RMSE} & \textbf{MAE} & \textbf{MAPE} & \textbf{MSE} & \textbf{RMSE} & \textbf{MAE} & \textbf{MAPE} & \textbf{MSE} & \textbf{RMSE} & \textbf{MAE} & \textbf{MAPE} & \textbf{MSE} & \textbf{RMSE} & \textbf{MAE} & \textbf{MAPE} \\ \hline
\multirow{2}{*}{\textbf{DePlot}}     & Normal                                     & \textbf{0.0181}       & \textbf{0.1214 }       & \textbf{0.1148}       & \textbf{33.1477}       & 0.0137       & 0.0924        & 0.0896       & 41.7180       & 0.0164       & 0.1131        & 0.1042       & 24.1590       & 0.0115       & 0.0780        & 0.0737       & 60.7436       \\
                                            & CoT                                        & 0.0479       & 0.1610        & 0.1553       & 68.0959       & \textbf{0.0068}       & \textbf{0.0699}        & \textbf{0.0643}       & \textbf{28.2178}       & \textbf{0.0150}       & \textbf{0.1038}        & \textbf{0.0963}       & \textbf{22.9814}       & \textbf{0.0050}       & \textbf{0.0624}        & \textbf{0.0562}      &\textbf{ 30.1001}       \\ \hline
\multirow{2}{*}{\textbf{Deepseek-vl2}} & Normal                                     & 0.0180       & 0.1111        & 0.1059       & 26.4934       & 0.0181       & 0.1012        & 0.0979       & 45.0789       & 0.0219       & 0.1191        & 0.1137       & 39.6341       & 0.0083       & 0.0719        & 0.0674       & 20.2569       \\
                                            & CoT                                        & \textbf{0.0129}       & \textbf{0.0910}        & \textbf{0.0846}       & \textbf{22.9444}       & \textbf{0.0067}       & \textbf{0.0635}        & \textbf{0.0595}       & \textbf{23.7390}       & \textbf{0.0086}       & \textbf{0.0717}        & \textbf{0.0674}       & \textbf{13.5817}       & \textbf{0.0054 }      & \textbf{0.0647}        & \textbf{0.0592}       & \textbf{21.8665}       \\ \hline
\multirow{2}{*}{\textbf{LLaVA}}             & Normal                                     &  \textbf{0.0046}        & \textbf{0.0618}         & \textbf{0.0481}        & \textbf{16.5818}         & \textbf{0.0010}        & \textbf{0.0288}         & \textbf{0.0265}        & \textbf{6.4364}         & \textbf{0.0072}         & \textbf{0.0770}  & \textbf{0.0711}        & \textbf{11.0233}        &  \textbf{0.0015}        & \textbf{0.0362}         & \textbf{0.0293}        & \textbf{3.8947}      \\
                                            & CoT                                        &  0.0050         &  0.0696          & 0.0597         & 18.0747         & 0.0021        &  0.0418          & 0.0361       & 8.6570         & 0.0100         & 0.0890          & 0.0815        & 14.1880        &  0.0060        & 0.0719          & 0.0631       & 8.2430        \\ \hline
\multirow{2}{*}{\textbf{Gemma}}             & Normal                                     & 0.0058        & 0.0644        & 0.0590        & 15.4665         & 0.0031        & 0.0467         & 0.0424        & 11.9728         & 0.0041        & 0.0561         & 0.0491        & 10.6249         & 0.0007        & 0.0232         & 0.0207        & 17.4091         \\
                                            & CoT                                        & \textbf{0.0047}        & \textbf{0.0552}         & \textbf{0.0498}        & \textbf{12.5399}         & \textbf{0.0026}        & \textbf{0.0387}         & \textbf{0.0348}        & \textbf{9.3983}         & \textbf{0.0030}        & \textbf{0.0488}         & \textbf{0.0405}        & \textbf{9.3557}         & \textbf{0.0006}        & \textbf{0.0222}         & \textbf{0.0201}        & \textbf{16.0692}         \\ \hline
\multirow{2}{*}{\textbf{Phi3}}              & Normal                                     &0.0095         & \textbf{0.0809}          & 0.0752        & \textbf{13.9770}         & 0.0014       & 0.0331          & 0.0306       & \textbf{12.1948}       & 0.0060        & 0.0713         &0.0659        & 16.1038         & 0.0018       & 0.0335         & 0.0305        & 19.0512           \\
                                            & CoT                                        &  \textbf{0.0087}       & 0.0815          & \textbf{0.0733}         & 14.5882           & \textbf{0.0007}          & \textbf{0.0254}        & \textbf{0.0223}          & 12.6890        &  \textbf{0.0023}        &  \textbf{0.0436}         & \textbf{0.0397}    & \textbf{7.9716}       & \textbf{0.0008}         & \textbf{0.0249}        &  \textbf{0.0195} 
                                            & \textbf{13.2321}
                                            \\ \hline        
\end{tabular}%
}
\caption{Comparison of predictive performance across four companies (Accor, Air Liquide, BNP Paribas, and Capgemini) using various vision-language models (Google DePlot, Deepseek-vl2, LLaVA, Gemma, and Phi3) under both normal and chain-of-thought (CoT) prompting. Metrics reported include MSE, RMSE, MAE, and MAPE (\%).
}
\label{tab:my-table-cot}
\end{table*}

To evaluate the effectiveness of Chain-of-Thought (CoT) prompting versus direct (normal) prompting, we assess five vision-language models (DePlot, Deepseek-vl2, LLaVA, Gemma, and Phi3) on stock price prediction tasks across four companies: Accor, Air Liquide, BNP Paribas, and Capgemini. As shown in Table \ref{tab:my-table-cot}, CoT prompting leads to improved predictive accuracy in most cases, as evidenced by lower MSE, MAE, and MAPE values.

For DePlot, normal prompting achieves lower MSEs across most companies, particularly on Accor (0.0181 vs. 0.0479) . However, CoT prompting improves performance on Air Liquide (0.0137 vs 0.0068) and Capgemini (0.0115 vs. 0.0050), suggesting case-dependent gains.
In contrast, Deepseek-vl2 benefits more consistently from CoT prompting. On Capgemini, MSE improves from 0.0083 to 0.0054 (a 34.94\% reduction), and on BNP Paribas from 0.0219 to 0.0086. So, the overall trend across the evaluated stocks favors CoT prompting.
LLaVA shows mixed results. CoT prompting leads to higher MSEs on all four companies. For example, on Accor, MSE increases from 0.0046 (normal) to 0.0050 (CoT), and on BNP Paribas from 0.0072 to 0.0100. Also on Capgemini CoT does not show an improvement(0.0015 vs. 0.0060), indicating that LLaVA may not benefit from reasoning-style prompting in stock prediction.
For Gemma, CoT prompting results in consistent improvements across all companies. On Accor, the MSE decreases from 0.0058 to 0.0047 (a 19\% improvement), and similar trends are seen on Air Liquide (0.0031 to 0.0026) and Capgemini (0.0007 to 0.0006). This suggests that Gemma’s performance benefits from explicit reasoning during prediction.
Phi3 also shows favorable gains with CoT prompting. On Accor, MSE improves from 0.0095 to 0.0087, and on Capgemini from 0.0018 to 0.0008—a notable 55.56\% improvement. Air Liquide and BNP Paribas also show reduced MSEs under CoT, reinforcing the advantage of step-by-step reasoning for this model.

All in all, Chain-of-Thought prompting enhances predictive performance for most VLMs and stocks, with particularly strong improvements observed in Deepseek-vl2, Gemma, and Phi3. While not universally better, CoT prompting proves to be an effective strategy in many scenarios, especially when paired with models capable of leveraging the added reasoning structure.

\subsubsection{comparison to the ARIMA baseline}
\begin{figure}[h]
    \centering
    \includegraphics[width=0.45\textwidth]{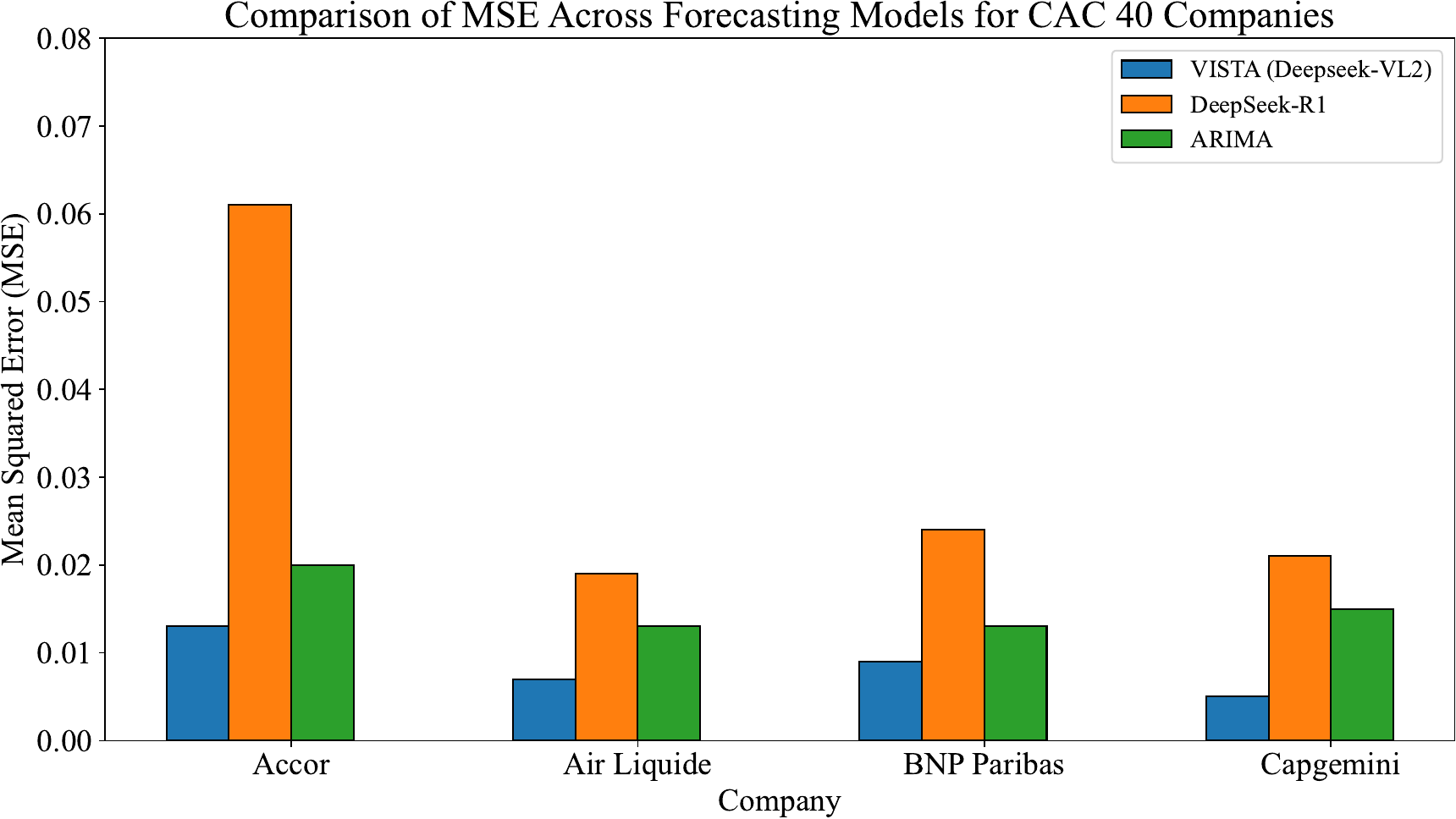}
    \caption{Performance of the prediction methods compared to ARIMA baseline.}
    \label{fig:bar-plot}
\end{figure}

To evaluate the effectiveness of VISTA against a well-established baseline, we compare it with the AutoRegressive Integrated Moving Average (ARIMA) model \cite{box2015time} using four datasets selected from Yahoo Finance. For each dataset, we use an observation window of 100 stock values to predict the subsequent 5 values, and compute the mean squared error (MSE) with respect to the ground truth. We then calculate the average MSE across all segments within each dataset and present the results in Fig. \ref{fig:bar-plot}. Regarding the results, DeepSeek‑R1 is a general‑purpose language model whose weights were optimized for open‑domain reasoning, code, etc. It has no time‑series inductive bias. ARIMA, in contrast, is purpose‑built for univariate auto‑correlated signals; it fits a few coefficients directly to the stock’s recent history. When data are limited and the horizon is short, that narrow specialization often beats a large but unfocused model. This pattern has been documented repeatedly in the forecasting literature, e.g. \cite{makridakis2018m4}, where classical methods outperformed ML and deep nets on a large benchmark of univariate series. This accounts for the lower predictive performance of DeepSeek-R1 compared to ARIMA. In contrast, our VISTA approach incorporates a visual representation of the stock price trajectory, enabling it to more effectively capture the nuanced variations within the time series. As a result, VISTA outperforms both the ARIMA model and the text-only DeepSeek-R1 approach.

\section{Ablation and Discussions}

\begin{table*}[]\label{tab:my-table-noise}
\centering
\resizebox{\textwidth}{!}{%
\begin{tabular}{lllllllllllll}
\hline
\multirow{2}{*}{\textbf{Stock Name}} & \multirow{2}{*}{\textbf{No Noise}} & \multicolumn{11}{c}{\textbf{Added Noise Coefficient}}                                                                                                                                                \\ \cline{3-13} 
                                     &                                    & \textbf{0.010} & \textbf{0.015} & \textbf{0.020} & \textbf{0.025} & \textbf{0.030} & \textbf{0.040} & \textbf{0.045} & \textbf{0.055} & \textbf{0.060} &
                                     \textbf{0.065} &
                                     \textbf{0.070} \\ \hline
\textbf{Accor}                       & 0.0569                             & 0.0600         & 0.0578         & 0.0658         & 0.0643         & 0.0624        & 0.0607         & 0.0668          & 0.0686         & 0.0695  & 0.0707 &  0.0888      \\
\textbf{CapGemini}                   &  0.0123                             & 0.0144         & 0.0120         & 0.0130         & 0.0126         & 0.0128        & 0.0135         & 0.0149       & 0.0155         & 0.0142 & 0.0157 &  0.0135       \\
\textbf{AirLiquide}                  &  0.0161                             & 0.0161         & 0.0175         & 0.0190         & 0.0177         &  0.0202         & 0.0193         & 0.0176                 & 0.0312         & 0.0163 & 0.0294 & 0.0318       \\ \hline
\end{tabular}%
}

\caption{MSE of DePlot on three stocks (Accor, Cap Gemini, Air Liquide) as a function of injected salt‑and‑pepper noise.
The left‑most column is the clean image baseline. Subsequent columns correspond to noise densities 0.010 → 0.070. A fixed salt‑vs‑pepper ratio of 0.2 and a deterministic seed (42) are used to ensure reproducibility.}
\label{tab:noise-added-table}
\end{table*}

Table \ref{tab:noise-added-table} quantifies how deliberately corrupting the input plot degrades forecasting accuracy. For each stock we kept the textual price history unchanged and progressively injected salt‑and‑pepper artifacts into the time-series plot before prompting DePlot. MSE rises monotonically with the noise coefficient: for Accor the MSE almost doubles from 0.0569 (clean) to 0.0888 at a 7\% corruption rate, while Cap Gemini and Air Liquide exhibit analogous—though less steep—deterioration. Because every other experimental variable (prompt wording, numerical values, decoding temperature = 0) is held constant, this trend isolates the visual channel as the decisive factor.

These findings substantiate our central claim that VISTA’s gains stem from true vision‑language fusion rather than from the language backbone alone. If the model were ignoring the image and relying purely on the serialized numbers, MSE would be invariant to visual perturbations; instead we observe a dose‑dependent penalty once salient chart structure is obscured. Put differently, the line graph supplies spatial cues (slope, extrema alignment, triangle formations, etc.) that the language‑only representation cannot convey, and VISTA exploits that extra signal. The ablation therefore offers evidence that high‑quality visual input is not merely decorative but materially improves short‑horizon stock forecasting, a motivation for multimodal inference in financial time‑series analysis. The visual context—whether from charts, financial tables, or annotated figures—appears to enhance model understanding of market dynamics. Moreover, the fact that these improvements are realized without architectural modifications beyond the vision input stream highlights the effectiveness of multimodal fusion for financial forecasting.


\subsection{CoT prompting}
Across all models, CoT prompting demonstrates the greatest benefit for lower-capacity or smaller-scale models, where structured reasoning may help compensate for limited model depth. In contrast, larger or vision-native models with strong direct prediction capabilities may not always benefit from CoT and, in some cases, perform better with simpler prompts.

These findings suggest that while CoT prompting enhances model performance in some cases, its effectiveness is model- and data-dependent. It offers significant gains in difficult or visually complex scenarios, but care must be taken when applying it universally, especially in well-structured or less ambiguous cases.
\section{Conclusion}
In this work, we introduced VISTA, a novel training-free framework that utilizes Vision-Language Models (VLMs) for multi-modal stock forecasting. Our evaluations across LLM-VLM pairs with comparable architectures demonstrate that incorporating visual representations of time-series data provides complementary signals that improve predictive accuracy beyond what is achievable using textual input alone. Additionally, we improved the quality of textual reasoning through the use of chain-of-thought (CoT) prompting, which consistently outperformed direct prediction prompts. Together, these findings demonstrate the effectiveness of VISTA in combining visual and textual modalities with structured reasoning to enable more accurate and insightful financial predictions.
{
    \small
    \bibliographystyle{unsrt}
    \bibliography{main}

@String(AAAI = {AAAI})

@article{chen2024visionts,
  title={Visionts: Visual masked autoencoders are free-lunch zero-shot time series forecasters},
  author={Chen, Mouxiang and Shen, Lefei and Li, Zhuo and Wang, Xiaoyun Joy and Sun, Jianling and Liu, Chenghao},
  journal={arXiv preprint arXiv:2408.17253},
  year={2024}
}

@inproceedings{zhou2021informer,
  title={Informer: Beyond efficient transformer for long sequence time-series forecasting},
  author={Zhou, Haoyi and Zhang, Shanghang and Peng, Jieqi and Zhang, Shuai and Li, Jianxin and Xiong, Hui and Zhang, Wancai},
  booktitle={Proceedings of the AAAI conference on artificial intelligence},
  volume={35},
  number={12},
  pages={11106--11115},
  year={2021}
}

@article{lim2021temporal,
  title={Temporal fusion transformers for interpretable multi-horizon time series forecasting},
  author={Lim, Bryan and Ar{\i}k, Sercan {\"O} and Loeff, Nicolas and Pfister, Tomas},
  journal={International Journal of Forecasting},
  volume={37},
  number={4},
  pages={1748--1764},
  year={2021},
  publisher={Elsevier}
}

@article{wu2021autoformer,
  title={Autoformer: Decomposition transformers with auto-correlation for long-term series forecasting},
  author={Wu, Haixu and Xu, Jiehui and Wang, Jianmin and Long, Mingsheng},
  journal={Advances in neural information processing systems},
  volume={34},
  pages={22419--22430},
  year={2021}
}

@inproceedings{he2022masked,
  title={Masked autoencoders are scalable vision learners},
  author={He, Kaiming and Chen, Xinlei and Xie, Saining and Li, Yanghao and Doll{\'a}r, Piotr and Girshick, Ross},
  booktitle={Proceedings of the IEEE/CVF conference on computer vision and pattern recognition},
  pages={16000--16009},
  year={2022}
}

@article{bao2021beit,
  title={Beit: Bert pre-training of image transformers},
  author={Bao, Hangbo and Dong, Li and Piao, Songhao and Wei, Furu},
  journal={arXiv preprint arXiv:2106.08254},
  year={2021}
}

@book{achelis2000technical,
  title={Technical Analysis from A to Z, 2nd Edition},
  author={Achelis, S.B.},
  isbn={9780071380119},
  url={https://books.google.dj/books?id=XuiF-2eWHYUC},
  year={2000},
  publisher={McGraw Hill LLC}
}

@inproceedings{devlin2019bert,
  title={Bert: Pre-training of deep bidirectional transformers for language understanding},
  author={Devlin, Jacob and Chang, Ming-Wei and Lee, Kenton and Toutanova, Kristina},
  booktitle={Proceedings of the 2019 conference of the North American chapter of the association for computational linguistics: human language technologies, volume 1 (long and short papers)},
  pages={4171--4186},
  year={2019}
}

@article{beyer2024paligemma,
  title={Paligemma: A versatile 3b vlm for transfer},
  author={Beyer, Lucas and Steiner, Andreas and Pinto, Andr{\'e} Susano and Kolesnikov, Alexander and Wang, Xiao and Salz, Daniel and Neumann, Maxim and Alabdulmohsin, Ibrahim and Tschannen, Michael and Bugliarello, Emanuele and others},
  journal={arXiv preprint arXiv:2407.07726},
  year={2024}
}

@inproceedings{radford2021learning,
  title={Learning transferable visual models from natural language supervision},
  author={Radford, Alec and Kim, Jong Wook and Hallacy, Chris and Ramesh, Aditya and Goh, Gabriel and Agarwal, Sandhini and Sastry, Girish and Askell, Amanda and Mishkin, Pamela and Clark, Jack and others},
  booktitle={International conference on machine learning},
  pages={8748--8763},
  year={2021},
  organization={PmLR}
}

@misc{li2022blipbootstrappinglanguageimagepretraining,
      title={BLIP: Bootstrapping Language-Image Pre-training for Unified Vision-Language Understanding and Generation}, 
      author={Junnan Li and Dongxu Li and Caiming Xiong and Steven Hoi},
      year={2022},
      eprint={2201.12086},
      archivePrefix={arXiv},
      primaryClass={cs.CV},
      url={https://arxiv.org/abs/2201.12086}, 
}

@misc{liu2023deplotoneshotvisuallanguage,
      title={DePlot: One-shot visual language reasoning by plot-to-table translation}, 
      author={Fangyu Liu and Julian Martin Eisenschlos and Francesco Piccinno and Syrine Krichene and Chenxi Pang and Kenton Lee and Mandar Joshi and Wenhu Chen and Nigel Collier and Yasemin Altun},
      year={2023},
      eprint={2212.10505},
      archivePrefix={arXiv},
      primaryClass={cs.CL},
      url={https://arxiv.org/abs/2212.10505}, 
}

@misc{zhai2023sigmoidlosslanguageimage,
      title={Sigmoid Loss for Language Image Pre-Training}, 
      author={Xiaohua Zhai and Basil Mustafa and Alexander Kolesnikov and Lucas Beyer},
      year={2023},
      eprint={2303.15343},
      archivePrefix={arXiv},
      primaryClass={cs.CV},
      url={https://arxiv.org/abs/2303.15343}, 
}

@misc{gemmateam2024gemmaopenmodelsbased,
      title={Gemma: Open Models Based on Gemini Research and Technology}, 
      author={Gemma Team and Thomas Mesnard and Cassidy Hardin and Robert Dadashi and Surya Bhupatiraju and Shreya Pathak and Laurent Sifre and Morgane Rivière and Mihir Sanjay Kale and Juliette Love and Pouya Tafti and Léonard Hussenot and Pier Giuseppe Sessa and Aakanksha Chowdhery and Adam Roberts and Aditya Barua and Alex Botev and Alex Castro-Ros and Ambrose Slone and Amélie Héliou and Andrea Tacchetti and Anna Bulanova and Antonia Paterson and Beth Tsai and Bobak Shahriari and Charline Le Lan and Christopher A. Choquette-Choo and Clément Crepy and Daniel Cer and Daphne Ippolito and David Reid and Elena Buchatskaya and Eric Ni and Eric Noland and Geng Yan and George Tucker and George-Christian Muraru and Grigory Rozhdestvenskiy and Henryk Michalewski and Ian Tenney and Ivan Grishchenko and Jacob Austin and James Keeling and Jane Labanowski and Jean-Baptiste Lespiau and Jeff Stanway and Jenny Brennan and Jeremy Chen and Johan Ferret and Justin Chiu and Justin Mao-Jones and Katherine Lee and Kathy Yu and Katie Millican and Lars Lowe Sjoesund and Lisa Lee and Lucas Dixon and Machel Reid and Maciej Mikuła and Mateo Wirth and Michael Sharman and Nikolai Chinaev and Nithum Thain and Olivier Bachem and Oscar Chang and Oscar Wahltinez and Paige Bailey and Paul Michel and Petko Yotov and Rahma Chaabouni and Ramona Comanescu and Reena Jana and Rohan Anil and Ross McIlroy and Ruibo Liu and Ryan Mullins and Samuel L Smith and Sebastian Borgeaud and Sertan Girgin and Sholto Douglas and Shree Pandya and Siamak Shakeri and Soham De and Ted Klimenko and Tom Hennigan and Vlad Feinberg and Wojciech Stokowiec and Yu-hui Chen and Zafarali Ahmed and Zhitao Gong and Tris Warkentin and Ludovic Peran and Minh Giang and Clément Farabet and Oriol Vinyals and Jeff Dean and Koray Kavukcuoglu and Demis Hassabis and Zoubin Ghahramani and Douglas Eck and Joelle Barral and Fernando Pereira and Eli Collins and Armand Joulin and Noah Fiedel and Evan Senter and Alek Andreev and Kathleen Kenealy},
      year={2024},
      eprint={2403.08295},
      archivePrefix={arXiv},
      primaryClass={cs.CL},
      url={https://arxiv.org/abs/2403.08295}, 
}

@misc{yahoo_finance,
  title = {Yahoo Finance},
  howpublished = {\url{https://finance.yahoo.com}},
  note = {Accessed: 2025-02-01}
}

@misc{euronext_paris,
  title = {Euronext Paris},
  author = {{Euronext}},
  howpublished = {\url{https://www.euronext.com/en/markets/paris}},
  note = {Accessed: 2025-02-01}
}

@book{dehaene2003three,
  author    = {Stanislas Dehaene and Elizabeth Spelke and Stanislas Pinel and Jean-Pierre Stanescu and Elizabeth Tsivkin},
  title     = {Three parietal circuits for number processing},
  journal   = {Cognitive Neuropsychology},
  volume    = {20},
  number    = {3},
  pages     = {487--506},
  year      = {2003},
  publisher = {Taylor \& Francis}
}

@article{fuster2009cortex,
  author  = {Joaqu{\'\i}n M. Fuster},
  title   = {Cortex and memory: emergence of a new paradigm},
  journal = {Journal of Cognitive Neuroscience},
  volume  = {21},
  number  = {11},
  pages   = {2047--2072},
  year    = {2009}
}

@book{kosslyn2006graph,
  author    = {Stephen M. Kosslyn},
  title     = {Graph Design for the Eye and Mind},
  publisher = {Oxford University Press},
  year      = {2006}
}

@article{tversky2002animation,
  author  = {Barbara Tversky and Julie Morrison},
  title   = {Animation: Can it facilitate?},
  journal = {International Journal of Human-Computer Studies},
  volume  = {57},
  number  = {4},
  pages   = {247--262},
  year    = {2002}
}

@book{murphy1999technical,
  author    = {John J. Murphy},
  title     = {Technical Analysis of the Financial Markets: A Comprehensive Guide to Trading Methods and Applications},
  publisher = {New York Institute of Finance},
  year      = {1999}
}

@article{alayrac2022flamingo,
  author  = {Jean-Baptiste Alayrac and Jeff Donahue and Pauline Luc and Antoine Miech and Iain Barr and Yana Hasson and Karel Lenc and Arthur Mensch and Katie Millican and Malcolm Reynolds and Roman Ring and Eliza Rutherford and Serkan Cabi and Tengda Han and Mohammad Babaeizadeh and Shakir Mohamed and João Carreira and Lucas Smaira and Oriol Vinyals and Andrew Zisserman and Relja Arandjelović},
  title   = {Flamingo: a Visual Language Model for Few-Shot Learning},
  journal = {arXiv preprint arXiv:2204.14198},
  year    = {2022}
}

@article{openai2023gpt4,
  author  = {OpenAI},
  title   = {GPT-4 Technical Report},
  journal = {arXiv preprint arXiv:2303.08774},
  year    = {2023}
}

@article{stockwell1996localization,
  title={Localization of the complex spectrum: the S transform},
  author={Stockwell, Robert G. and Mansinha, Lalu and Lowe, Richard P.},
  journal={IEEE Transactions on Signal Processing},
  volume={44},
  number={4},
  pages={998--1001},
  year={1996},
  publisher={IEEE}
}

@article{wei2022chain,
  title={Chain-of-thought prompting elicits reasoning in large language models},
  author={Wei, Jason and Wang, Xuezhi and Schuurmans, Dale and Bosma, Maarten and Xia, Fei and Chi, Ed and Le, Quoc V and Zhou, Denny and others},
  journal={Advances in neural information processing systems},
  volume={35},
  pages={24824--24837},
  year={2022}
}

@article{2020t5,
  author  = {Colin Raffel and Noam Shazeer and Adam Roberts and Katherine Lee and Sharan Narang and Michael Matena and Yanqi Zhou and Wei Li and Peter J. Liu},
  title   = {Exploring the Limits of Transfer Learning with a Unified Text-to-Text Transformer},
  journal = {Journal of Machine Learning Research},
  year    = {2020},
  volume  = {21},
  number  = {140},
  pages   = {1-67},
  url     = {http://jmlr.org/papers/v21/20-074.html}
}

@article{raffel2020exploring,
  title={Exploring the limits of transfer learning with a unified text-to-text transformer},
  author={Raffel, Colin and Shazeer, Noam and Roberts, Adam and Lee, Katherine and Narang, Sharan and Matena, Michael and Zhou, Yanqi and Li, Wei and Liu, Peter J},
  journal={Journal of machine learning research},
  volume={21},
  number={140},
  pages={1--67},
  year={2020}
}

@misc{liu2022deplot,
      title={DePlot: One-shot visual language reasoning by plot-to-table translation},
      author={Liu, Fangyu and Eisenschlos, Julian Martin and Piccinno, Francesco and Krichene, Syrine and Pang, Chenxi and Lee, Kenton and Joshi, Mandar and Chen, Wenhu and Collier, Nigel and Altun, Yasemin},
      year={2022},
      eprint={2212.10505},
      archivePrefix={arXiv},
      primaryClass={cs.CL}
}

@misc{meta_llama_3.1_8b_instruct_2024,
  author       = {Meta AI},
  title        = {Llama 3.1 8B Instruct},
  year         = {2024},
  url          = {https://huggingface.co/meta-llama/Llama-3.1-8B-Instruct},
  publisher    = {Hugging Face}
}

@misc{meta_llava-1.5-7b-hf_2024,
  author       = {Meta AI},
  title        = {LLaVA-v1.5-7B},
  year         = {2024},
  url          = {https://huggingface.co/llava-hf/llava-1.5-7b-hf},
  publisher    = {Hugging Face}
}

@article{touvron2023llama,
  title={Llama: Open and efficient foundation language models},
  author={Touvron, Hugo and Lavril, Thibaut and Izacard, Gautier and Martinet, Xavier and Lachaux, Marie-Anne and Lacroix, Timoth{\'e}e and Rozi{\`e}re, Baptiste and Goyal, Naman and Hambro, Eric and Azhar, Faisal and others},
  journal={arXiv preprint arXiv:2302.13971},
  year={2023}
}

@article{liu2023visual,
  title={Visual instruction tuning},
  author={Liu, Haotian and Li, Chunyuan and Wu, Qingyang and Lee, Yong Jae},
  journal={Advances in neural information processing systems},
  volume={36},
  pages={34892--34916},
  year={2023}
}

@article{abdin2024phi,
  title={Phi-3 technical report: A highly capable language model locally on your phone},
  author={Abdin, Marah and Aneja, Jyoti and Awadalla, Hany and Awadallah, Ahmed and Awan, Ammar Ahmad and Bach, Nguyen and Bahree, Amit and Bakhtiari, Arash and Bao, Jianmin and Behl, Harkirat and others},
  journal={arXiv preprint arXiv:2404.14219},
  year={2024}
}

@misc{microsoft_phi3_mini_128k_instruct_2024,
  author       = {Microsoft},
  title        = {Phi-3 Mini 128K Instruct},
  year         = {2024},
  publisher    = {Hugging Face},
  url          = {https://huggingface.co/microsoft/Phi-3-mini-128k-instruct}
}

@misc{microsoft_phi3_vision_128k_instruct_2024,
  author       = {Microsoft},
  title        = {Phi-3 Vision 128K Instruct},
  year         = {2024},
  publisher    = {Hugging Face},
  url          = {https://huggingface.co/microsoft/Phi-3-vision-128k-instruct}
}

@article{team2024gemma,
  title={Gemma: Open models based on gemini research and technology},
  author={Team, Gemma and Mesnard, Thomas and Hardin, Cassidy and Dadashi, Robert and Bhupatiraju, Surya and Pathak, Shreya and Sifre, Laurent and Rivi{\`e}re, Morgane and Kale, Mihir Sanjay and Love, Juliette and others},
  journal={arXiv preprint arXiv:2403.08295},
  year={2024}
}

@article{gemma_2025,
    title={Gemma 3},
    url={https://goo.gle/Gemma3Report},
    publisher={Kaggle},
    author={Gemma Team},
    year={2025}
}

@article{gemma3_technical_report_2024,
  title   = {Gemma 3 Technical Report},
  author  = {Google DeepMind},
  journal = {Google AI},
  year    = {2024},
  url     = {https://ai.google.dev/gemma/docs/core}
}

@article{lu2024deepseek,
  title={Deepseek-vl: towards real-world vision-language understanding},
  author={Lu, Haoyu and Liu, Wen and Zhang, Bo and Wang, Bingxuan and Dong, Kai and Liu, Bo and Sun, Jingxiang and Ren, Tongzheng and Li, Zhuoshu and Yang, Hao and others},
  journal={arXiv preprint arXiv:2403.05525},
  year={2024}
}

@article{guo2025deepseek,
  title={Deepseek-r1: Incentivizing reasoning capability in llms via reinforcement learning},
  author={Guo, Daya and Yang, Dejian and Zhang, Haowei and Song, Junxiao and Zhang, Ruoyu and Xu, Runxin and Zhu, Qihao and Ma, Shirong and Wang, Peiyi and Bi, Xiao and others},
  journal={arXiv preprint arXiv:2501.12948},
  year={2025}
}

@misc{wu2024deepseekvl2mixtureofexpertsvisionlanguagemodels,
      title={DeepSeek-VL2: Mixture-of-Experts Vision-Language Models for Advanced Multimodal Understanding}, 
      author={Zhiyu Wu and Xiaokang Chen and Zizheng Pan and Xingchao Liu and Wen Liu and Damai Dai and Huazuo Gao and Yiyang Ma and Chengyue Wu and Bingxuan Wang and Zhenda Xie and Yu Wu and Kai Hu and Jiawei Wang and Yaofeng Sun and Yukun Li and Yishi Piao and Kang Guan and Aixin Liu and Xin Xie and Yuxiang You and Kai Dong and Xingkai Yu and Haowei Zhang and Liang Zhao and Yisong Wang and Chong Ruan},
      year={2024},
      eprint={2412.10302},
      archivePrefix={arXiv},
      primaryClass={cs.CV},
      url={https://arxiv.org/abs/2412.10302}, 
}

@misc{deepseekai2025deepseekr1incentivizingreasoningcapability,
      title={DeepSeek-R1: Incentivizing Reasoning Capability in LLMs via Reinforcement Learning}, 
      author={DeepSeek-AI},
      year={2025},
      eprint={2501.12948},
      archivePrefix={arXiv},
      primaryClass={cs.CL},
      url={https://arxiv.org/abs/2501.12948}, 
}

@book{box2015time,
  title={Time Series Analysis: Forecasting and Control},
  author={Box, George E. P. and Jenkins, Gwilym M. and Reinsel, Gregory C. and Ljung, Greta M.},
  edition={5th},
  year={2015},
  publisher={John Wiley \& Sons},
}

@article{makridakis2018m4,
  title        = {The M4 Competition: Results, Findings, Conclusion and Way Forward},
  author       = {Makridakis, Spyros and Spiliotis, Evangelos and Assimakopoulos, Vassilios},
  journal      = {International Journal of Forecasting},
  volume       = {34},
  number       = {4},
  pages        = {802--808},
  year         = {2018},
  doi          = {10.1016/j.ijforecast.2018.06.001}
}

@article{razmara2024fever,
  title={Fever detection with infrared thermography: Enhancing accuracy through machine learning techniques},
  author={Razmara, Parsa and Khezresmaeilzadeh, Tina and Jenkins, B Keith},
  journal={arXiv preprint arXiv:2407.15302},
  year={2024}
}

@article{khezresmaeilzadeh2025preserving,
  title={Preserving Privacy and Utility in LLM-Based Product Recommendations},
  author={Khezresmaeilzadeh, Tina and Zhang, Jiang and Andreadis, Dimitrios and Psounis, Konstantinos},
  journal={arXiv preprint arXiv:2505.00951},
  year={2025}
}

@inproceedings{azizi2025mambaextend,
  title={Mambaextend: A training-free approach to improve long context extension of mamba},
  author={Azizi, Seyedarmin and Kundu, Souvik and Sadeghi, Mohammad Erfan and Pedram, Massoud},
  booktitle={The Thirteenth International Conference on Learning Representations}
}

@article{kolyshkina2021interpretability,
  title={Interpretability of machine learning solutions in public healthcare: The CRISP-ML approach},
  author={Kolyshkina, Inna and Simoff, Simeon},
  journal={Frontiers in big data},
  volume={4},
  pages={660206},
  year={2021},
  publisher={Frontiers Media SA}
}

@article{khanal2020systematic,
  title={A systematic review: machine learning based recommendation systems for e-learning},
  author={Khanal, Shristi Shakya and Prasad, PWC and Alsadoon, Abeer and Maag, Angelika},
  journal={Education and Information Technologies},
  volume={25},
  number={4},
  pages={2635--2664},
  year={2020},
  publisher={Springer}
}

@article{vaswani2017attention,
  title={Attention is all you need},
  author={Vaswani, Ashish and Shazeer, Noam and Parmar, Niki and Uszkoreit, Jakob and Jones, Llion and Gomez, Aidan N and Kaiser, {\L}ukasz and Polosukhin, Illia},
  journal={Advances in neural information processing systems},
  volume={30},
  year={2017}
}

@misc{dosovitskiy2021imageworth16x16words,
      title={An Image is Worth 16x16 Words: Transformers for Image Recognition at Scale}, 
      author={Alexey Dosovitskiy and Lucas Beyer and Alexander Kolesnikov and Dirk Weissenborn and Xiaohua Zhai and Thomas Unterthiner and Mostafa Dehghani and Matthias Minderer and Georg Heigold and Sylvain Gelly and Jakob Uszkoreit and Neil Houlsby},
      year={2021},
      eprint={2010.11929},
      archivePrefix={arXiv},
      primaryClass={cs.CV},
      url={https://arxiv.org/abs/2010.11929}, 
}

@misc{fayyazi2025facterfairnessawareconformalthresholding,
      title={FACTER: Fairness-Aware Conformal Thresholding and Prompt Engineering for Enabling Fair LLM-Based Recommender Systems}, 
      author={Arya Fayyazi and Mehdi Kamal and Massoud Pedram},
      year={2025},
      eprint={2502.02966},
      archivePrefix={arXiv},
      primaryClass={cs.IR},
      url={https://arxiv.org/abs/2502.02966}, 
}

@misc{strzelecki2022machine,
  title={Machine learning for biomedical application},
  author={Strzelecki, Micha{\l} and Badura, Pawel},
  journal={Applied Sciences},
  volume={12},
  number={4},
  pages={2022},
  year={2022},
  publisher={MDPI}
}

@inproceedings{abdollahi2025icd,
  title={ICD 2 S: A Hybrid Ising-Classical-Machines Data-Driven QUBO Solver Method},
  author={Abdollahi, Armin and Kamal, Mehdi and Pedram, Massoud},
  booktitle={Proceedings of the 30th Asia and South Pacific Design Automation Conference},
  pages={914--920},
  year={2025}
}

@article{khezresmaeilzadeh2024echoes,
  title={Echoes of Privacy: Uncovering the Profiling Practices of Voice Assistants},
  author={Khezresmaeilzadeh, Tina and Zhu, Elaine and Grieco, Kiersten and Dubois, Daniel J and Psounis, Konstantinos and Choffnes, David},
  journal={arXiv preprint arXiv:2409.07444},
  year={2024}
}

@INPROCEEDINGS{8406613,
  author={Papernot, Nicolas and McDaniel, Patrick and Sinha, Arunesh and Wellman, Michael P.},
  booktitle={2018 IEEE European Symposium on Security and Privacy (EuroS\&P)}, 
  title={SoK: Security and Privacy in Machine Learning}, 
  year={2018},
  volume={},
  number={},
  pages={399-414},
  doi={10.1109/EuroSP.2018.00035}}

@ARTICLE{9433648,
  author={De Cristofaro, Emiliano},
  journal={IEEE Security \& Privacy}, 
  title={A Critical Overview of Privacy in Machine Learning}, 
  year={2021},
  volume={19},
  number={4},
  pages={19-27},
  doi={10.1109/MSEC.2021.3076443}}

@inproceedings{sadeghi2024peano,
  title={Peano-vit: Power-efficient approximations of non-linearities in vision transformers},
  author={Sadeghi, Mohammad Erfan and Fayyazi, Arash and Azizi, Seyedarmin and Pedram, Massoud},
  booktitle={Proceedings of the 29th ACM/IEEE International Symposium on Low Power Electronics and Design},
  pages={1--6},
  year={2024}
}

@INPROCEEDINGS{10179281,
  author={Salem, Ahmed and Cherubin, Giovanni and Evans, David and Köpf, Boris and Paverd, Andrew and Suri, Anshuman and Tople, Shruti and Zanella-Béguelin, Santiago},
  booktitle={2023 IEEE Symposium on Security and Privacy (SP)}, 
  title={SoK: Let the Privacy Games Begin! A Unified Treatment of Data Inference Privacy in Machine Learning}, 
  year={2023},
  volume={},
  number={},
  pages={327-345},
  doi={10.1109/SP46215.2023.10179281}}

@article{gu2023mamba,
  title={Mamba: Linear-time sequence modeling with selective state spaces},
  author={Gu, Albert and Dao, Tri},
  journal={arXiv preprint arXiv:2312.00752},
  year={2023}
}

@article{portugal2018use,
  title={The use of machine learning algorithms in recommender systems: A systematic review},
  author={Portugal, Ivens and Alencar, Paulo and Cowan, Donald},
  journal={Expert Systems with Applications},
  volume={97},
  pages={205--227},
  year={2018},
  publisher={Elsevier}
}

@article{li2021survey,
  title={A survey of convolutional neural networks: analysis, applications, and prospects},
  author={Li, Zewen and Liu, Fan and Yang, Wenjie and Peng, Shouheng and Zhou, Jun},
  journal={IEEE transactions on neural networks and learning systems},
  volume={33},
  number={12},
  pages={6999--7019},
  year={2021},
  publisher={IEEE}
}

@article{gogas2021machine,
  title={Machine learning in economics and finance},
  author={Gogas, Periklis and Papadimitriou, Theophilos},
  journal={Computational Economics},
  volume={57},
  pages={1--4},
  year={2021},
  publisher={Springer}
}

@article{ranjbar2025beyond,
  title={Beyond Subjective Measures: Systematic Review of Deep Learning in Chronic Pain: Modalities, Methods, and Applications},
  author={Ranjbar, Melika Ahmadi and Ghaleh, Alireza and Dogaheh, Habib Bakian and Razmara, Parsa and Baghani, Matin},
  year={2025}
}

@inproceedings{abbasi2024fedgreen,
  title={FedGreen: Carbon-aware Federated Learning with Model Size Adaptation},
  author={Abbasi, Ali and Dong, Fan and Wang, Xin and Leung, Henry and Zhou, Jiayu and Drew, Steve},
  booktitle={2024 IEEE International Conference on Communications Workshops (ICC Workshops)},
  pages={1352--1358},
  year={2024},
  organization={IEEE}
}

@inproceedings{aziziqmambaextend,
  title={QMambaExtend: Improving Long-Context Extension of Memory-Efficient Mamba Models},
  author={Azizi, Seyedarmin and Kundu, Souvik and Sadeghi, Mohammad Erfan and Pedram, Massoud},
  booktitle={First Workshop on Scalable Optimization for Efficient and Adaptive Foundation Models}
}
}


\end{document}